# Data Fusion of Deep Learned Molecular Embeddings for Property Prediction

Robert J. Appleton[1], Brian C. Barnes[2], Alejandro Strachan[1*]

[1] School of Materials Engineering and Birck Nanotechnology Center, Purdue University, West Lafayette, Indiana 47907, USA

[2] U.S. Army Combat Capabilities Development Command Army Research Laboratory, Aberdeen Proving Ground, Maryland 21005, USA

## Abstract

Data-driven approaches such as deep learning can result in predictive models for material properties with exceptional accuracy and efficiency. However, in many applications, data is sparse, severely limiting their accuracy and applicability. To improve predictions, techniques such as transfer learning and multi-task learning have been used. The performance of multi-task learning models depends on the strength of the underlying correlations between tasks and the completeness of the dataset. Standard multi-task models tend to underperform when trained on sparse datasets with weakly correlated properties. To address this gap, we fuse deep-learned embeddings generated by independent pre-trained single-task models, resulting in a multi-task model that inherits rich, property-specific representations. By re-using (rather than re-training) these embeddings, the resulting fused model outperforms standard multi-task models and can be extended with fewer trainable parameters. We demonstrate this technique on a widely used benchmark dataset of quantum chemistry data for small molecules as well as a newly compiled sparse dataset of experimental data collected from literature and our own quantum chemistry and thermochemical calculations.

* Corresponding author: strachan@purdue.edu

## Introduction

Materials discovery and design is crucial to the development of novel technologies that push the capabilities of science and engineering. For any specific targeted discovery effort, the process of identifying new candidate materials requires consideration of several properties to properly classify the use of a new material[1]. Typically, novel synthesis and subsequent experimental testing is costly and time-consuming, and thus there is a need for predictive methods to help characterize theoretical materials prior to synthesis and experimentation. In many areas of materials science, physical models are lacking and typically come with a strong tradeoff between accuracy and computational expense[2]. The use of machine learning in materials science has proven successful at accelerating this process in various ways such as material property prediction models[3–21], machine learning interatomic potentials[22–35], novel structure generation[9,36–40], and capturing microstructural effects to material response[41–44]. The types of models range from simpler models such as multiple linear regression models[45–47] (MLR), decision trees[48] (DT), random forests[49] (RF), and multi-layer perceptrons[50] (MLP), to more advanced models such as deep neural networks (DNN), convolutional neural networks[51] (CNN), graph neural networks[52–55] (GNN), such as graph convolutional networks (GCN) and message-passing neural networks[3,4,56,57] (MPNN), and transformer models[58] (used in large-language models (LLMs)). The model architecture implemented in this work is a directed MPNN developed within the *chemprop*[3,4] framework which is particularly designed for operating on molecules and molecular materials. A summary of the different parts of the models created in *chemprop* is described in a later section, but the important thing to note about this model is that it learns molecular/material properties from the molecular graph in a deep learning fashion.

In many materials applications, datasets are sparse and thus not suited for the application of typical deep learning techniques. Methods such as transfer learning[6–8,10,11] can help improve predictability on a data-poor target by leveraging learned information corresponding to a data-rich target (ideally related). This has been demonstrated in previous works using energetic materials data and the *chemprop* framework[10,11]. In this work, we show that data fusion techniques can be used to combine the deep-learned information of various single-task models and used to train a multi-task model with the entire dataset. We explore the performance of this technique with two large molecular datasets. We compile a large but very sparse dataset of properties spanning experimental measurements of crystal density, thermal stability and sensitivity as well as performing our own quantum chemistry and thermochemical calculations. To test the performance of this technique on a complete dataset, we use a common benchmark for model development of small molecules strongly correlated properties from quantum chemistry calculations. The results show that the fused, multi-task models outperform standard multi-task models for both the sparse dataset and the complete dataset while also providing enhanced prediction on several properties compared to single-task models.

Multimodal learning refers to the integration of information from distinct data types [59] and has been recently used in materials science, for example in predictive modeling of Li-ion solid electrolytes [60] and classification of 3D X-ray tomography data [61]. In molecular property prediction, Lu *et al.* [62] demonstrated that fusing traditional molecular features (SMILES-encoded vectors and extended connectivity fingerprints) with and molecular graph representations resulted in improved predictive accuracy, reliability, and noise resistance.

Our work is conceptually related but distinct. Instead of combining fundamentally different data modalities, we fuse molecular embeddings, each derived from message-passing neural networks trained independently for different target properties. This creates a scenario closer to intra-modal fusion: although the embeddings originate from the same underlying graph modality, they encode complementary chemical and structural information because each was optimized under a unique task. By unifying these embeddings within a multi-task framework, we aim to capture richer, property-specific representations while retaining the benefits of shared learning across tasks. In this sense, our approach extends the spirit of multimodal learning to the fusion of specialized deep learned embeddings within a common molecular graph domain.

## Methods

### Data

In this work, we curated a dataset of ~30K CHNOFCl molecules that sample the chemical space relevant to energetic materials (EMs) in a well distributed manner. Initially, we obtain ~127K molecules from various experimental property datasets[63–65] and a subset of Pubchem[66,67] with high oxygen balance ($OB_{100} > -60$). Oxygen balance is a widely used heuristic in the EM community and describes the amount of oxygen relative to the amount of carbon and hydrogen, as shown in its definition here:

$$OB_{100} \equiv \frac{100}{n_{atoms}\left(n_O - 2n_C - \frac{n_H}{2}\right)} \tag{1}$$

This expression is derived from the idea that one source of energy released by an energetic is the oxidation of carbon and hydrogen forming products such as $CO_2$ and $H_2O$[68]. Energetics typically have $OB_{100}$ near zero, while nearly all the molecules in the subset of Pubchem are normally distributed around -100, see Figure S1 of the Supplemental Material, indicating that most of these molecules are “oxygen-poor” or “fuel-rich”. Starting from this initial set of ~127K molecules, we implemented a multivariate bucket selection scheme to ensure the molecules were well distributed across various molecular characteristics. Specifically, we looked at the 2D distributions between oxygen balance (OB) and molecular weight (MW), OB and nitrogen percentage (N%), and MW and N% (see Figure S2(a)). The goal of the selection scheme is to make the sizes of the buckets more uniformly distributed in each 2D distribution. To do this, we first sorted the molecules in each bucket in order of molecular similarity to a reference dataset of EMs. The similarity metric is a number from 0 to 1 defined by computing the Tanimoto similarity[69] between the Morgan Fingerprints[70] (radius of 5) of two molecules. For each molecule in our dataset, we compute the similarity with respect to each molecule in a reference dataset of EMs[12,71–75] and take the maximum value. Now that the molecules are ordered by similarity, we then set a maximum value, $k$, on the number of molecules that can be in a single bucket. The top $k$ most similar molecules are selected and if a bucket has more than $k$ molecules the remaining are discarded. We performed this on our data first using a $k$ value of 225 on the 2D distribution of OB and MW and again using a $k$ value of 97 on the 2-d distribution of MW and N%. This results in only ~20K molecules being selected and the corresponding distributions of OB, MW, and N% are shown in Figure S2(b). As shown in Figure S2(b), the selected molecules are much more evenly distributed about the different molecular characteristics considered. This selection process resulted in a set of molecules that well sample chemical space based on their composition and size but did not consider the underlying molecular structure. It is obvious that molecules can be in similar compositional space but vastly different chemical space due to the variety of different ways the same composition can arrange itself in molecular structure. We compared the distribution of different EM-relevant substructures for our selected ~20K molecules with the reference dataset of EM molecules and found that several EM-relevant substructures were currently underrepresented, see Figure S3. To address this, we sampled molecules from two other existing sources, a subset of a theoretical dataset generated by PNNL[76] selected by similarity to EMs[11] and a subset of known CHNOFCl molecules from scientific literature obtained from the ChEMBL dataset[77]. The sampling was done to specifically target molecules that contain the various underrepresented substructures and resulting in an additional ~10K molecules. More details about the data sampling are shown in the Supplemental Material. The final distribution of EM-relevant substructures shows a good representation across the different substructures considered, see Figure S3. To ensure the distribution of our characteristics described above (OB, MW, and N%) did not get corrupted by this addition of molecules we observed the 2D distribution in Figure S2(c) and find that though the distributions are slightly altered they do not contain any drastic biases. We believe this final dataset of ~30K molecules well samples chemical space based on our analysis of the distribution of different composition-based characteristics and the distribution of EM-relevant substructures.

Using the selected molecules, we query open-source materials property datasets[63–65] and collect data from the EM literature[12,72–75] targeting experimental measurements related to thermal stability (melting temperature ($T_{melt}$) and decomposition temperature ($T_{dec}$)) and safety (impact sensitivity (IS) and friction sensitivity (FS)). We also targeted crystal density ($\rho_0$) due to its abundance and its relevance to energetic properties such as detonation performance. We note that for any molecule with multiple reports for the same measurement, the median was used for training. We removed 12 $T_{melt}$ data points from the dataset where the standard deviation across multiple measurements was greater than 50 K[64]. The $\rho_0$ data[63,65] all had standard deviations less than 0.1 g/cc. IS and FS exhibit significant variability, spanning four orders of magnitude ($10^{-1}$-$10^{2}$). To address this, we apply $\log_{10}$ transformation in line with previous works[11,12,16,78]. From this search we obtain the following amounts of data for each property shown in Table 1.

**Table 1.** Summary of experimental property data obtained from open literature.

| Property (exp) | # Datapoints |
|---|---|
| $T_{melt}$ (K)[64,72] | 3934 |
| $T_{dec}$ (K)[72,74] | 738 |
| $\rho_0$ (g/cc)[63,65] | 11582 |
| IS (J)[12,72,73,75] | 846 |
| FS (N)[72] | 274 |

For the non-halogen containing molecules (no F or Cl), we utilize a physics-based workflow to generate calculated molecular, crystalline, and detonation properties. Starting from the SMILES string, the python package *RDKit*[79] is used to generate several 3D configurations of conformers and compute their corresponding energies with the MMFF94 classical force field to determine the lowest energy conformer. The 3D structure of the lowest energy conformer is then used for various quantum chemistry calculations via the density functional theory (DFT) code *Gaussian 16*[80] at the B3LYP/6-31G** level. Using the outputs of these DFT calculations (3D electron density and electrostatic potential point grids), a quantitative structure property relationship (QSPR)[81–84] model is used to estimate the crystalline density ($\rho_0$) and crystalline heat of formation ($\Delta H^0_f$). From these estimates, we can now utilize the thermochemical code *Cheetah*[85] to solve for the Chapman-Jouget (C-J) detonation conditions: explosive energy ($E_{expl}$), detonation velocity ($V_{det}$), detonation pressure ($P_{det}$), and detonation temperature ($T_{det}$). This method has been widely used over the past two decades for estimating crystalline properties[81–84] and generating high-throughput detonation data for ML models[13,14]. We collect these properties for over 23K molecules. Through this process we have compiled a large dataset of molecules that well sample chemical space as well as various experimental (when available) and calculated properties. The resulting dataset is very sparse as depicted by the varying amount of available experimental data for each property of interest in Table 1.

To compare the proposed model performance on a complete dataset we also utilize the QM9 dataset[86]. This dataset is a commonly used benchmark[3,4,23,24,87–91] for model development of small organic molecules and contains ~134k molecules with 12 properties that are calculated using quantum chemistry methods. We suspected that the fused, multi-task directed message passing neural network (F-MT D-MPNN) will not see the same level of improvement on this dataset compared to the sparse dataset we compiled above. However, we find that the F-MT D-MPNN model can maintain numerically comparable accuracy or better compared to the single-task directed message passing neural network (ST D-MPNN) on several properties, while the standard MT D-MPNN struggles to maintain the same level of accuracy. These results demonstrate the robustness of our method.

## Model Design

As stated above, the models developed in this work are built using the *chemprop*[3,4] framework. A model created with *chemprop* takes a 2-dimensional graph representation of the molecule derived from the atomic connectivity described in the SMILES string. In a molecular graph, the atoms are treated as nodes and the bonds are treated like edges. Information containing atomic features and bond features is propagated through the graph in a directed fashion using a directed MPNN with various trainable parameters, see Figure 1(a). The output of the MPNN is aggregated and flattened to a 1-dimensional vector that is referred to as the molecular embedding. The molecular embedding is then passed to a feed-forward neural network (FFN) which outputs the prediction. A more detailed description of these types of models is available in the following references[3,4]. A powerful aspect of this model is that the parameters for both the MPNN and FFN are optimized simultaneously during training, and the molecular embedding is specifically learned for the property (or properties, see Figure 1(b)) being considered. One can imagine having multiple properties of interest and training multiple ST D-MPNN models in *chemprop* that each learn a different set of model parameters resulting in a unique and specifically tailored molecular embedding for each property. On the other hand, one can also train a single model in *chemprop* that tries to learn the properties from the same model and thus share a single "global" molecular embedding from which all the properties are predicted (see Figure 1(b)). The results of such MT D-MPNN can in theory produce enhanced accuracy (with respect to a ST model) on all properties due to the inductive transfer learning from co-training. However, there have been studies that show a variety of different results[3,16,92–97], and the actual benefits of MT modelling really depend on the underlying relationships between each task, careful construction of a loss function with contributions from each task, and the size and completeness of the dataset. In this work, we aim to improve MT modelling capabilities within *chemprop* by implementing a data fusion approach to combine the molecular embeddings of ST models that can be ingested by a multi-head FFN. This new approach is shown in Figure 1(c) and will be referred to as a F-MT D-MPNN. In this work, we compare the results of training the F-MT D-MPNN with two data fusion techniques: (1) naïve concatenation of the molecular embeddings, and (2) performing principal component analysis (PCA) on the concatenated molecular embedding.

To evaluate the performance of a F-MT D-MPNN on a sparse dataset, we compare the accuracy across the 13 different properties in our compiled dataset between the standard ST and MT D-MPNNs in *chemprop* (Figure 1(a,b)) with the proposed F-MT D-MPNN (Figure 1(c)). To enable a robust evaluation of model performance on unseen data, we employ a nested 5-fold cross-validation scheme. Hyperparameter optimization (HPO) is performed within the inner folds using only the training split of the inner fold, leveraging the Hyperopt search algorithm[98] alongside the Asynchronous Successive Halving Algorithm (ASHA) scheduler[99]. The inner fold models are evaluated using the testing split of the inner fold and the best-performing hyperparameters are then used to retrain models on the training split of the outer fold. Final performance is assessed on the outer test split, which remains completely unseen during both HPO and model retraining. All data splits are shared across all models. Models were trained until the validation loss fails to improve for 20 consecutive epochs, with early stopping typically triggered within 200 epochs across all experiments. The optimized and trained ST D-MPNNs were used to generate the molecular embeddings that are used by the F-MT D-MPNN. We compare a F-MT D-MPNN trained on all 13 properties as well as 5 other F-MT D-MPNNs that are trained on a subset of related properties. The subsets include thermal stability ($T_{melt}$ and $T_{dec}$), crystal properties ($\rho_0$ and $\Delta H^0_f$), detonation properties ($E_{expl}$, $V_{det}$, $P_{det}$, and $T_{det}$), sensitivity measurements (IS and FS), and molecular properties ($E_{gap}$ and $\mu$). Similarly, we evaluate the performance of ST, MT, and F-MT D-MPNNs on the QM9 dataset to provide a benchmark on a complete dataset with strongly correlated properties.

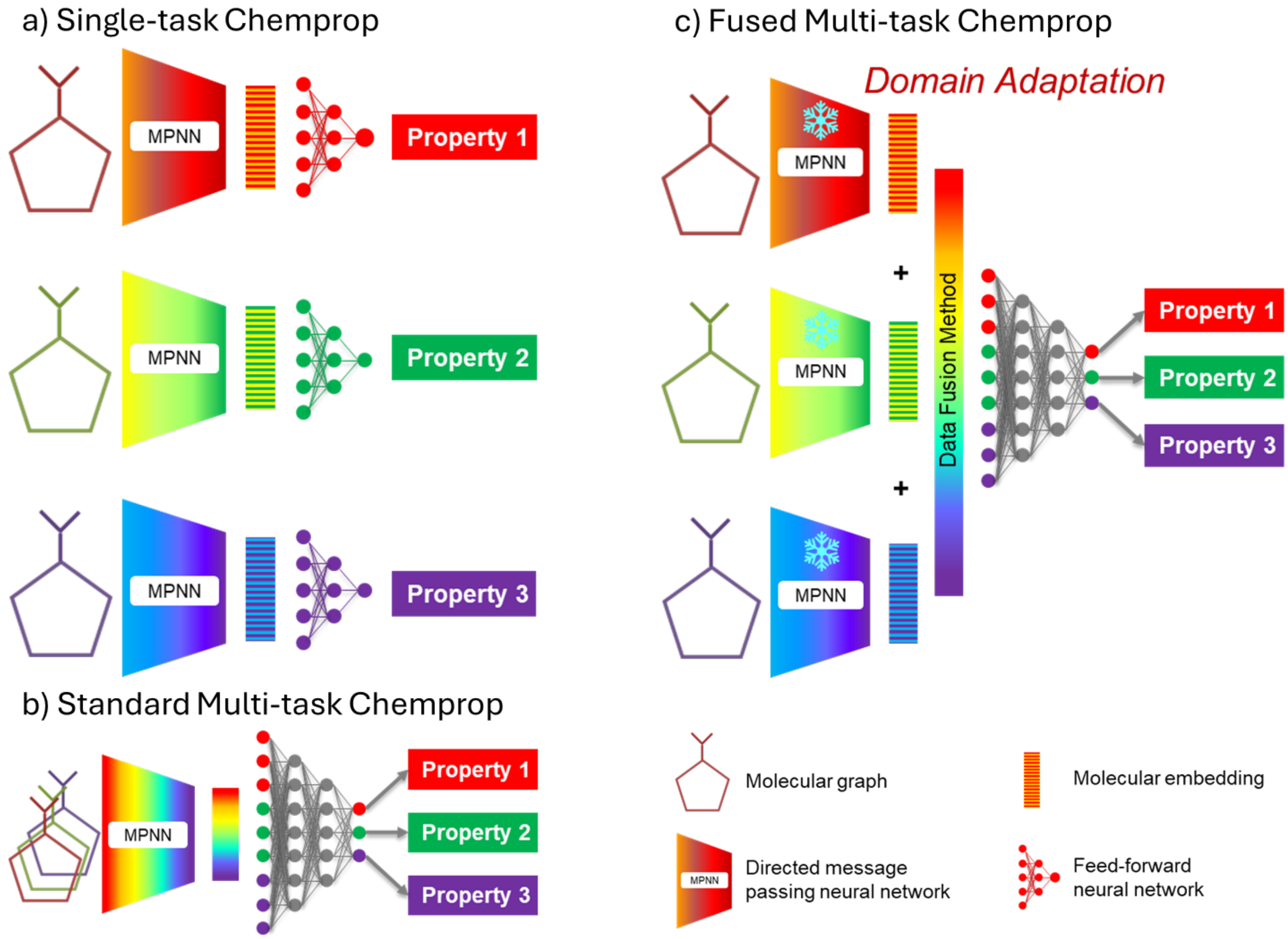

**Figure 1.** Diagrams of different architectures explored. Snow symbol indicates frozen weights after single-task training.

# Results

## Model Performance on a Sparse Dataset

We first discuss the performance of the F-MT model trained on all properties of the sparse dataset. Table 2 contains the test metrics for the ST, MT, and F-MT models for each property of interest (MT models trained on all 13 properties). Based the mean RMSE values across the 5 cross validation folds, the F-MT model performs best on 6 out of the 13 properties, the ST model performs best on 6 out of the 13 properties, and the MT model exhibits the lowest error in only 2 out of the 13 properties (ST and F-MT have equal mean RMSE for $E_{expl}$). Considering the standard deviation of the RMSE across the 5 cross validation splits, the F-MT model overlaps with the ST model on 9 out of the 13 properties and outperforms the MT model on $\rho_{calc}$, $\Delta H^0{}_f$, $V_{det}$, $P_{det}$, and $E_{gap}$, highlighting the improvements of F-MT over MT on data-rich properties. The ST model is clearly better on $\rho_{calc}$, $\Delta H^0{}_f$, $E_{gap}$, and μ. We note that MT learning proves most impactful for the properties where data is most limited ($T_{melt}$, $T_{dec}$, IS, and FS). This is expected given that these deep learning models typically require large amounts of data. To help visualize the impact of the F-MT models we have plotted the RMSE of each normalized so that the ST RMSE is 1 in Figure 2.

**Table 2.** Summary of test metrics across 5-fold cross validation comparing ST, MT, and F-MT models on the various properties of interest. The MT models in this table are trained on all 12 properties. The RMSE and $R^2$ values represent the mean across the 5-folds, and the standard deviation is shown in parentheses.

| Test Predictions | | Models | | | | | |
|---|---|---|---|---|---|---|---|
| Subsets | Properties of Interest | ST D-MPNN | | MT D-MPNN | | F-MT D-MPNN | |
| | | RMSE | $R^2$ | RMSE | $R^2$ | RMSE | $R^2$ |
| Thermal stability | $T_{melt}$ exp (K) | $37.1^{(0.9)}$ | $0.78^{(0.01)}$ | $37.1^{(0.6)}$ | $0.78^{(0.01)}$ | $36.7^{(1.6)}$ | $0.78^{(0.01)}$ |
| | $T_{dec}$ exp (K) | $53.6^{(5.3)}$ | $0.58^{(0.08)}$ | $48.4^{(4.3)}$ | $0.66^{(0.06)}$ | $49.3^{(4.2)}$ | $0.65^{(0.06)}$ |
| Crystal Properties | $\rho_0$ exp (g/cc) | $0.061^{(0.005)}$ | $0.87^{(0.02)}$ | $0.066^{(0.002)}$ | $0.85^{(0.01)}$ | $0.064^{(0.003)}$ | $0.86^{(0.01)}$ |
| | $\rho_0$ calc (g/cc) | $0.020^{(0.001)}$ | $0.98^{(0.01)}$ | $0.027^{(0.001)}$ | $0.97^{(0.01)}$ | $0.023^{(0.001)}$ | $0.98^{(0.01)}$ |
| | $\Delta H^0{}_f$ calc (kcal/mol) | $7.3^{(0.4)}$ | $1.00^{(0.01)}$ | $14.7^{(1.0)}$ | $0.98^{(0.01)}$ | $11.4^{(1.1)}$ | $0.99^{(0.01)}$ |
| Detonation Properties | $E_{expl}$ calc (kJ/cc) | $0.44^{(0.03)}$ | $0.97^{(0.01)}$ | $0.49^{(0.03)}$ | $0.96^{(0.01)}$ | $0.44^{(0.03)}$ | $0.97^{(0.01)}$ |
| | $V_{det}$ calc (km/s) | $0.21^{(0.01)}$ | $0.97^{(0.01)}$ | $0.22^{(0.01)}$ | $0.96^{(0.01)}$ | $0.20^{(0.01)}$ | $0.97^{(0.01)}$ |
| | $P_{det}$ calc (GPa) | $1.11^{(0.06)}$ | $0.97^{(0.01)}$ | $1.25^{(0.06)}$ | $0.97^{(0.01)}$ | $1.10^{(0.04)}$ | $0.97^{(0.01)}$ |
| | $T_{det}$ calc (K) | $146.1^{(10.5)}$ | $0.96^{(0.01)}$ | $151.1^{(9.4)}$ | $0.96^{(0.01)}$ | $136.9^{(7.6)}$ | $0.97^{(0.01)}$ |
| Sensitivity Properties | IS exp (J)* | $0.35^{(0.03)}$ | $0.56^{(0.08)}$ | $0.35^{(0.02)}$ | $0.55^{(0.06)}$ | $0.34^{(0.01)}$ | $0.58^{(0.03)}$ |
| | FS exp (N)* | $0.45^{(0.06)}$ | $0.40^{(0.09)}$ | $0.41^{(0.09)}$ | $0.49^{(0.13)}$ | $0.42^{(0.02)}$ | $0.47^{(0.11)}$ |
| Molecular Properties | $E_{gap}$ calc (eV) | $0.30^{(0.01)}$ | $0.93^{(0.01)}$ | $0.36^{(0.01)}$ | $0.89^{(0.01)}$ | $0.33^{(0.02)}$ | $0.91^{(0.01)}$ |
| | μ calc (Debye) | $1.41^{(0.05)}$ | $0.62^{(0.03)}$ | $1.71^{(0.03)}$ | $0.53^{(0.02)}$ | $1.67^{(0.04)}$ | $0.55^{(0.02)}$ |

* The values for these properties were converted with $\log_{10}$ for training and so the RMSE does not correspond to the real units.

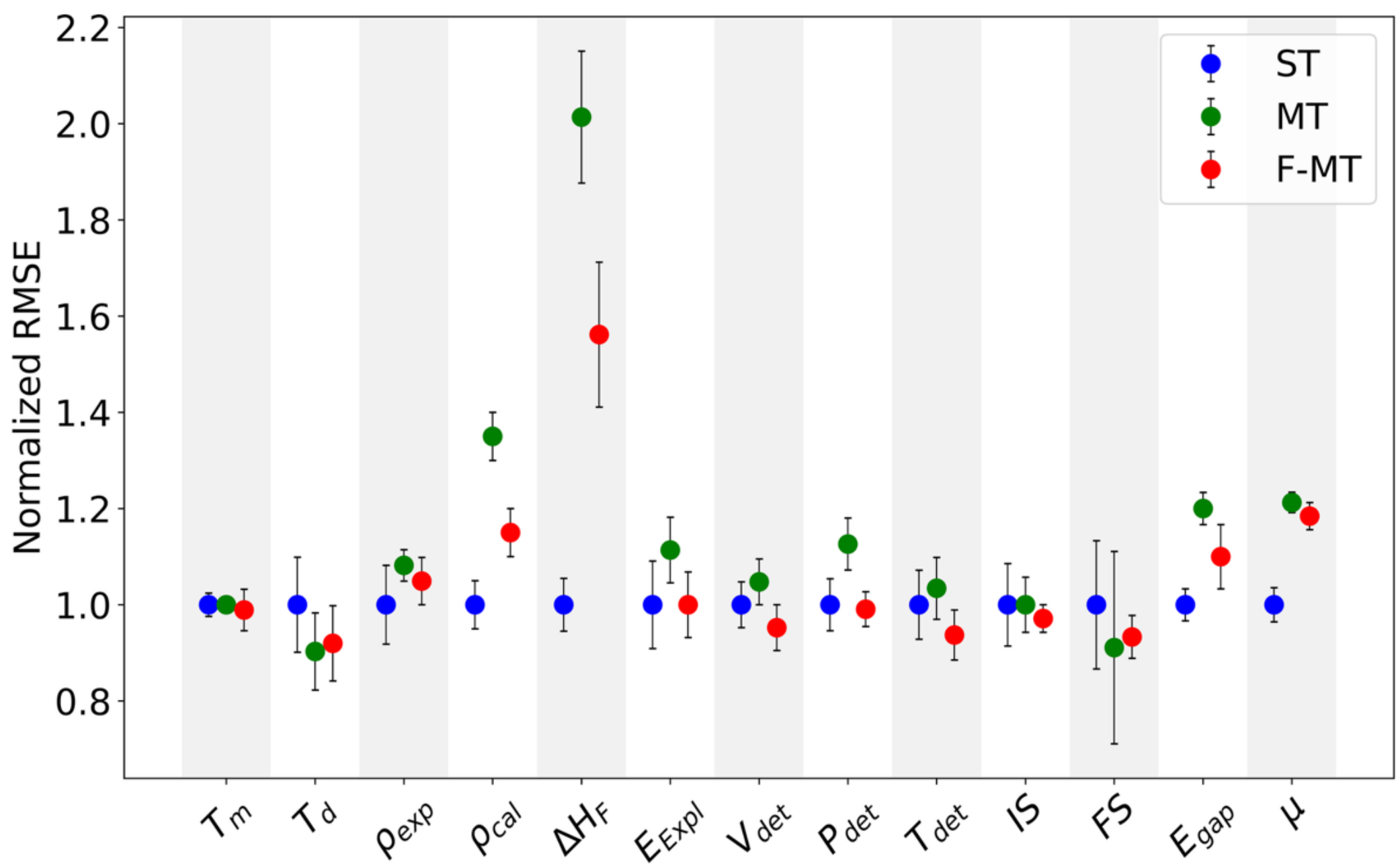

**Figure 2.** Normalized RMSE for each property of interest across the three model architectures. RMSE is normalized such that the ST RMSE is always 1. The error bars represent the standard deviation of the RMSE across the 5-folds. Points for same property are slightly horizontally offset to avoid overlapping error bars.

We now discuss the performance of the F-MT models trained on the 5 property subsets of the sparse dataset. Table 3 contains the test metrics for the ST, MT, and F-MT models for each property of interest (MT models trained on only the properties in each subset). Based on the mean RMSE values across the 5 cross validation folds, the F-MT model performs best on 8 out of the 13 properties, the ST model performs best on 5 out of the 13 properties, and the MT model exhibits the lowest error in only 2 out of the 13 properties (ST and F-MT have equal mean RMSE for $\rho_{exp}$ and $\rho_{calc}$). Considering the standard deviation of the RMSE across the 5 cross validation splits, the F-MT model and ST model overlap on 11 out of the 13 properties, where the F-MT model is clearly better on $V_{det}$ and the ST model is clearly better on μ. Moreover, the F-MT model outperforms the MT model on $\rho_{calc}$ and $\Delta H^0_f$, again highlighting the improvements of F-MT over MT on data-rich properties. The observation that MT learning proves most impactful for the properties where data is most limited is not as pronounced as what was seen when training on all 13 properties. The plotted the normalized RMSE of each property in Figure 3.

**Table 3.** Summary of test metrics across 5-fold cross validation comparing ST, MT, and F-MT models on the various properties of interest. The MT models in this table are trained on only the property subsets. The RMSE and $R^2$ values represent the mean across the 5-folds, and the standard deviation is shown in parentheses.

| Test Predictions | | Models | | | | | |
|---|---|---|---|---|---|---|---|
| Subsets | Properties of Interest | ST D-MPNN | | MT D-MPNN | | F-MT D-MPNN | |
| | | RMSE | $R^2$ | RMSE | $R^2$ | RMSE | $R^2$ |
| Thermal stability | $T_{melt}$ exp (K) | $37.1^{(0.9)}$ | $0.78^{(0.01)}$ | $37.6^{(1.2)}$ | $0.77^{(0.01)}$ | $36.3^{(1.7)}$ | $0.79^{(0.01)}$ |
| | $T_{dec}$ exp (K) | $53.6^{(5.3)}$ | $0.58^{(0.08)}$ | $51.0^{(3.0)}$ | $0.62^{(0.04)}$ | $52.3^{(6.0)}$ | $0.60^{(0.09)}$ |
| Crystal Properties | $\rho_0$ exp (g/cc) | $0.061^{(0.005)}$ | $0.87^{(0.02)}$ | $0.065^{(0.002)}$ | $0.86^{(0.01)}$ | $0.061^{(0.005)}$ | $0.87^{(0.02)}$ |
| | $\rho_0$ calc (g/cc) | $0.020^{(0.001)}$ | $0.98^{(0.01)}$ | $0.023^{(0.001)}$ | $0.98^{(0.01)}$ | $0.020^{(0.001)}$ | $0.98^{(0.01)}$ |
| | $\Delta H^0_f$ calc (kcal/mol) | $7.3^{(0.4)}$ | $1.00^{(0.01)}$ | $9.7^{(0.9)}$ | $0.99^{(0.01)}$ | $7.6^{(0.7)}$ | $1.00^{(0.01)}$ |
| Detonation Properties | $E_{expl}$ calc (kJ/cc) | $0.44^{(0.03)}$ | $0.97^{(0.01)}$ | $0.44^{(0.03)}$ | $0.97^{(0.01)}$ | $0.42^{(0.04)}$ | $0.97^{(0.01)}$ |
| | $V_{det}$ calc (km/s) | $0.21^{(0.01)}$ | $0.97^{(0.01)}$ | $0.20^{(0.01)}$ | $0.97^{(0.01)}$ | $0.19^{(0.01)}$ | $0.97^{(0.01)}$ |
| | $P_{det}$ calc (GPa) | $1.11^{(0.06)}$ | $0.97^{(0.01)}$ | $1.08^{(0.07)}$ | $0.97^{(0.01)}$ | $1.05^{(0.05)}$ | $0.98^{(0.01)}$ |
| | $T_{det}$ calc (K) | $146.1^{(10.5)}$ | $0.96^{(0.01)}$ | $135.7^{(11.8)}$ | $0.97^{(0.01)}$ | $127.7^{(12.3)}$ | $0.97^{(0.01)}$ |
| Sensitivity Properties | IS exp (J)* | $0.35^{(0.03)}$ | $0.56^{(0.08)}$ | $0.35^{(0.03)}$ | $0.54^{(0.07)}$ | $0.34^{(0.02)}$ | $0.57^{(0.07)}$ |
| | FS exp (N)* | $0.45^{(0.06)}$ | $0.40^{(0.09)}$ | $0.38^{(0.08)}$ | $0.55^{(0.15)}$ | $0.40^{(0.05)}$ | $0.50^{(0.06)}$ |
| Molecular Properties | $E_{gap}$ calc (eV) | $0.30^{(0.01)}$ | $0.93^{(0.01)}$ | $0.34^{(0.02)}$ | $0.91^{(0.01)}$ | $0.31^{(0.02)}$ | $0.92^{(0.01)}$ |
| | $\mu$ calc (Debye) | $1.41^{(0.05)}$ | $0.62^{(0.03)}$ | $1.61^{(0.02)}$ | $0.58^{(0.01)}$ | $1.63^{(0.03)}$ | $0.57^{(0.01)}$ |

* The values for these properties were converted with $\log_{10}$ for training and so the RMSE does not correspond to the real units.

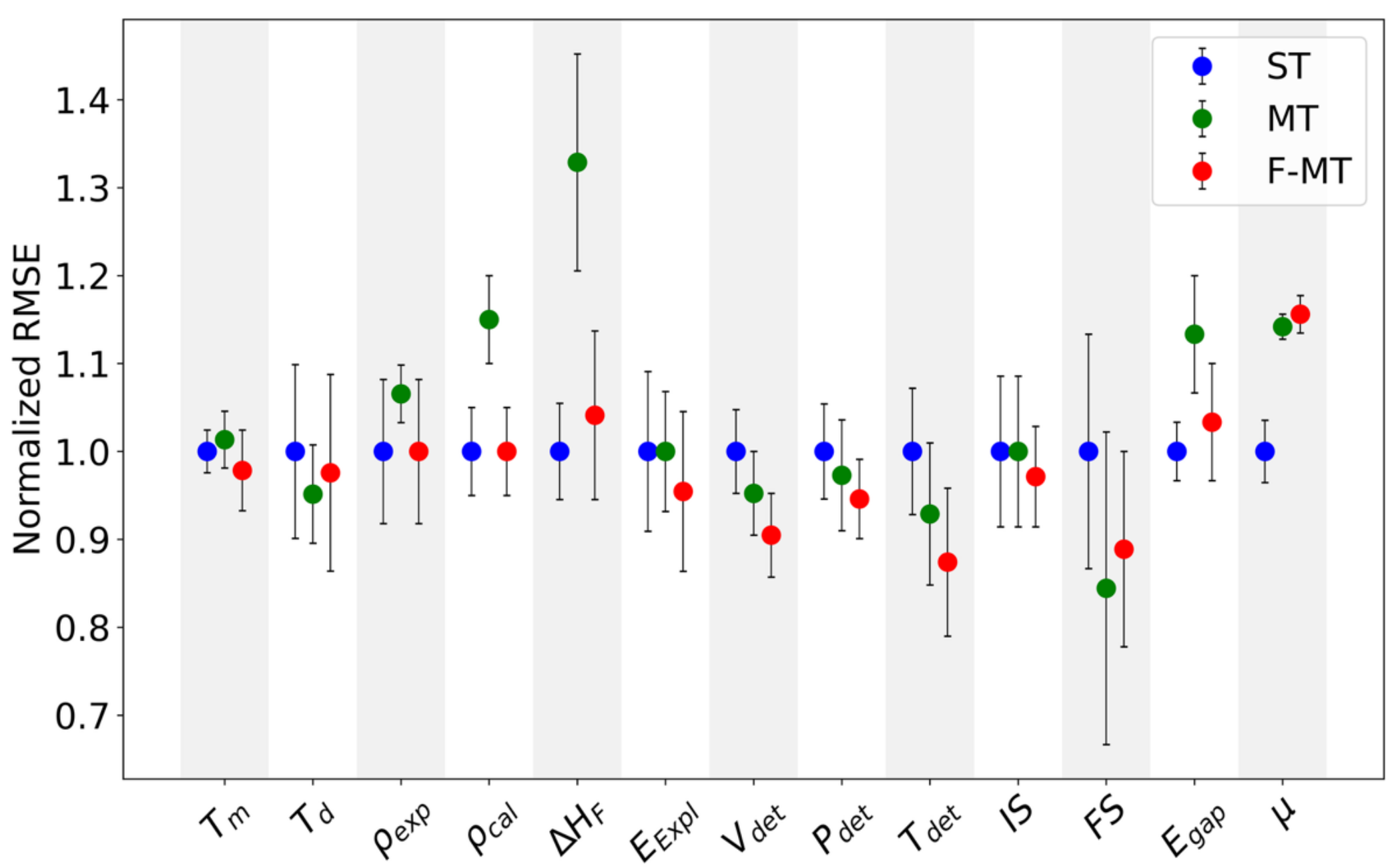

**Figure 3.** Normalized RMSE for each property of interest across the three model architectures. RMSE is normalized such that the ST RMSE is always 1. The error bars represent the standard deviation of the RMSE across the 5-folds.

Overall, these results demonstrate a clear advantage of F-MT models over the standard MT model. Both MT models (F-MT and standard MT) consistently outperform ST models on data-poor properties based on mean RMSE, owing to the benefits of co-training on data-rich and data-poor properties simultaneously. However, F-MT further improves upon standard MT by mitigating the loss in accuracy observed for data-rich properties. By preserving the deep learned molecular embeddings from each ST model, the F-MT architecture maintains high accuracy on well-represented properties while still gaining from shared learning. This advantage becomes even more pronounced when evaluating performance across different property subsets.

**Model Performance on a Complete Dataset**

In this section we discuss the performance of the F-MT model trained on all properties of the QM9 dataset. Table 4 contains the test metrics for the ST, MT, and F-MT model for each property of interest. We find that the F-MT model outperforms or is equivalent to the ST models on all 12 properties. Also, we find that the F-MT model outperforms or is equivalent to the standard MT model on 9 out of the 12 properties. The standard MT model does perform best on 3 of the properties (μ, r2, and ZPVE) but shows significant accuracy loss on 4 properties ($U_0$, $U_{298}$, $H_{298}$ and $G_{298}$). These results further emphasize the advantage of the F-MT model compared to the standard MT model. Furthermore, the mean absolute error (MAE) for $U_0$ is 0.24 +/- 0.01 Ha which would rank 12$^{th}$ on the leaderboard for this benchmark dataset[100].

**Table 4.** Summary of test metrics across 5-fold cross validation comparing ST, MT, and F-MT models on the various properties of interest. The RMSE and $R^2$ values represent the mean across the 5-folds and the standard deviation is shown in parentheses.

| Test Predictions | Models | | | | | |
|---|---|---|---|---|---|---|
| Properties of Interest | ST D-MPNN | | MT D-MPNN | | F-MT D-MPNN | |
| | RMSE | $R^2$ | RMSE | $R^2$ | RMSE | $R^2$ |
| μ (Debye) | $0.70^{(0.01)}$ | $0.79^{(0.01)}$ | $0.64^{(0.01)}$ | $0.83^{(0.01)}$ | $0.66^{(0.01)}$ | $0.81^{(0.01)}$ |
| α ($\alpha^3_0$) | $0.70^{(0.13)}$ | $0.99^{(0.01)}$ | $0.70^{(0.16)}$ | $0.99^{(0.01)}$ | $0.68^{(0.15)}$ | $0.99^{(0.01)}$ |
| HOMO (Ha) | $0.005^{(0.0001)}$ | $0.95^{(0.01)}$ | $0.005^{(0.0001)}$ | $0.95^{(0.01)}$ | $0.005^{(0.0001)}$ | $0.95^{(0.01)}$ |
| LUMO (Ha) | $0.005^{(0.0002)}$ | $0.99^{(0.01)}$ | $0.005^{(0.0002)}$ | $0.99^{(0.01)}$ | $0.005^{(0.0001)}$ | $0.99^{(0.01)}$ |
| $E_{gap}$ (Ha) | $0.007^{(0.0002)}$ | $0.98^{(0.01)}$ | $0.007^{(0.0004)}$ | $0.98^{(0.01)}$ | $0.007^{(0.0004)}$ | $0.98^{(0.01)}$ |
| r2 ($\alpha^2_0$) | $39.9^{(1.2)}$ | $0.98^{(0.01)}$ | $38.0^{(1.3)}$ | $0.98^{(0.01)}$ | $38.4^{(1.0)}$ | $0.98^{(0.01)}$ |
| ZPVE (Ha) | $0.005^{(0.0001)}$ | $1.00^{(0.01)}$ | $0.001^{(0.0001)}$ | $1.00^{(0.01)}$ | $0.005^{(0.0001)}$ | $1.00^{(0.01)}$ |
| $C_v$ (cal/(mol K)) | $0.27^{(0.02)}$ | $1.00^{(0.01)}$ | $0.28^{(0.01)}$ | $1.00^{(0.01)}$ | $0.26^{(0.01)}$ | $1.00^{(0.01)}$ |
| $U_0$ (Ha) | $0.46^{(0.06)}$ | $1.00^{(0.01)}$ | $0.69^{(0.06)}$ | $1.00^{(0.01)}$ | $0.34^{(0.04)}$ | $1.00^{(0.01)}$ |
| $U_{298}$ (Ha) | $0.41^{(0.12)}$ | $1.00^{(0.01)}$ | $0.69^{(0.06)}$ | $1.00^{(0.01)}$ | $0.34^{(0.04)}$ | $1.00^{(0.01)}$ |
| $H_{298}$ (Ha) | $0.64^{(0.21)}$ | $1.00^{(0.01)}$ | $0.69^{(0.06)}$ | $1.00^{(0.01)}$ | $0.34^{(0.04)}$ | $1.00^{(0.01)}$ |
| $G_{298}$ (Ha) | $0.55^{(0.09)}$ | $1.00^{(0.01)}$ | $0.69^{(0.06)}$ | $1.00^{(0.01)}$ | $0.34^{(0.04)}$ | $1.00^{(0.01)}$ |

**Exploring Connections Between Molecular Embeddings**

In this section, we explore the influence of the molecular embedding of each property on the predictions made by the F-MT model. CUR decomposition is a method for approximating a large matrix by using a selection of its actual columns, in contrast to other singular value decomposition (SVD) based methods that combine several columns of the original matrix to construct the components of the reduced matrix. By utilizing CUR, we can preserve the original meaning of each feature selected during decomposition. Applying CUR to compress the concatenated embedding down to only 50 features, we can observe how many features are selected from the molecular embedding of each property, see Figure 4.

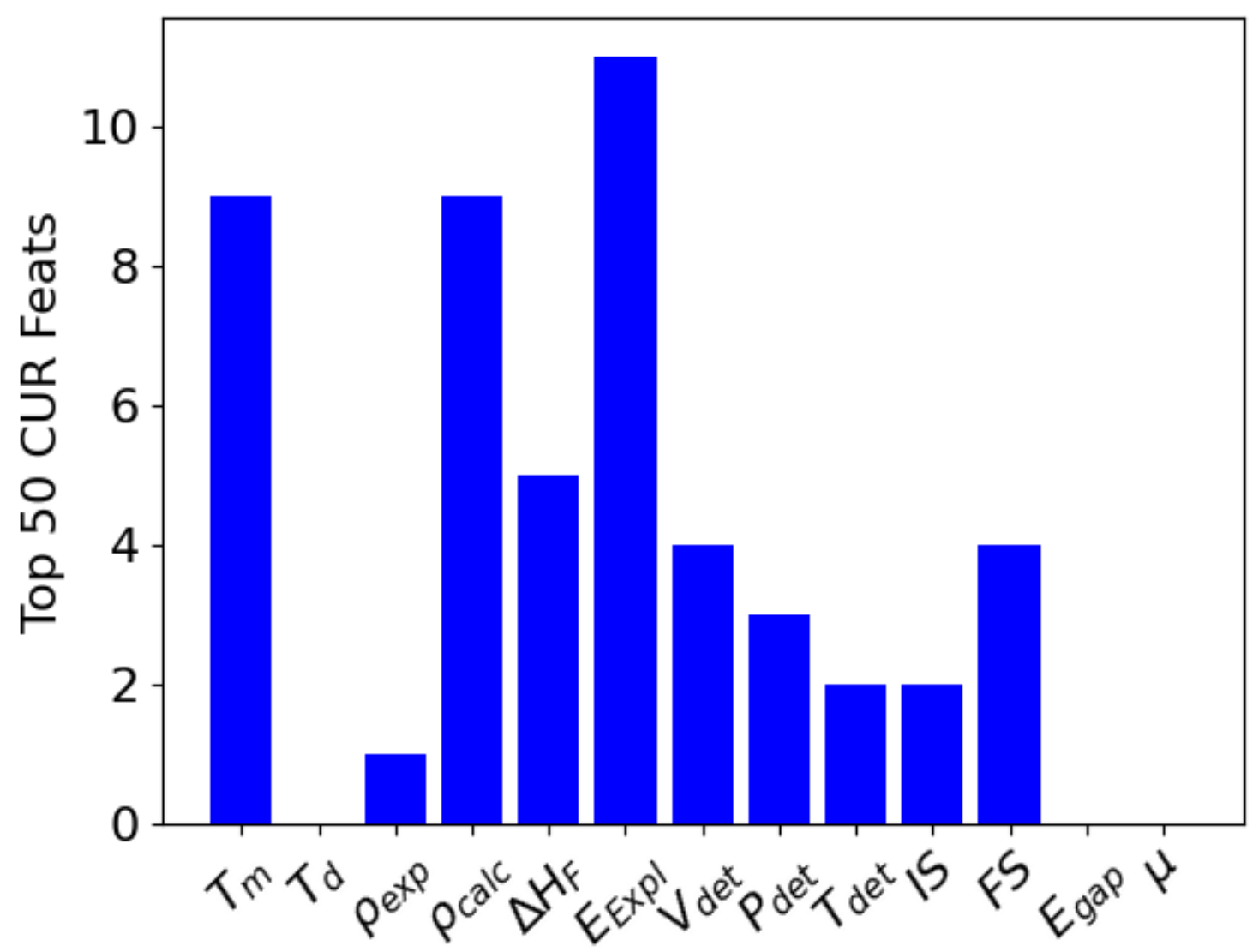

**Figure 4.** Number of features from the molecular embedding of each property selected by CUR to compress the concatenated embedding from 6000 to 50 features.

The CUR decomposition analysis allows us to see which embeddings contribute the most to the variability across the dataset, however it does not provide any intuition about how the embedding of one property influences the prediction on another property. Therefore, we propose a way to perform this analysis by systematically adding noise to one of the molecular embeddings and observing the changes in the accuracy for predicting the other properties. This is done by taking one of the molecular embeddings ($L_i$), introducing an amount of gaussian noise ($n$), and then evaluating the model predictions with the trained F-MT model. We then quantify the effect of $L_i$ on a given property ($P_j$) by computing the following ratio:

$$a_{ij} = \frac{RMSE_j(no\ noise)}{RMSE_j\ (L_i\ noise)},$$

where the $RMSE_j(no\ noise)$ is the RMSE on property $P_j$ without noise added to the $L_i$, and $RMSE_j\ (L_i\ noise)$ is the RMSE on property $P_j$ when noise has been added to $L_i$. Values of $a_{ij}$ that are close to 1 show that the $L_i$ does not have a strong effect on the prediction of $P_j$. Values of $a_{ij}$ that are much less than 1 show that the $L_i$ does influence the prediction of $P_j$. The gaussian noise is systematically generated for each molecular embedding by sampling random numbers from a normal distribution centered at 0 with a standard deviation 3 times the standard deviation of the values in the molecular embedding ($\sigma_{L_i}$). The noise is then added to $L_i$ and predictions are made with the trained F-MT model. The $a_{ij}$ values for applying noise to each molecular embedding is displayed in Figure 5.

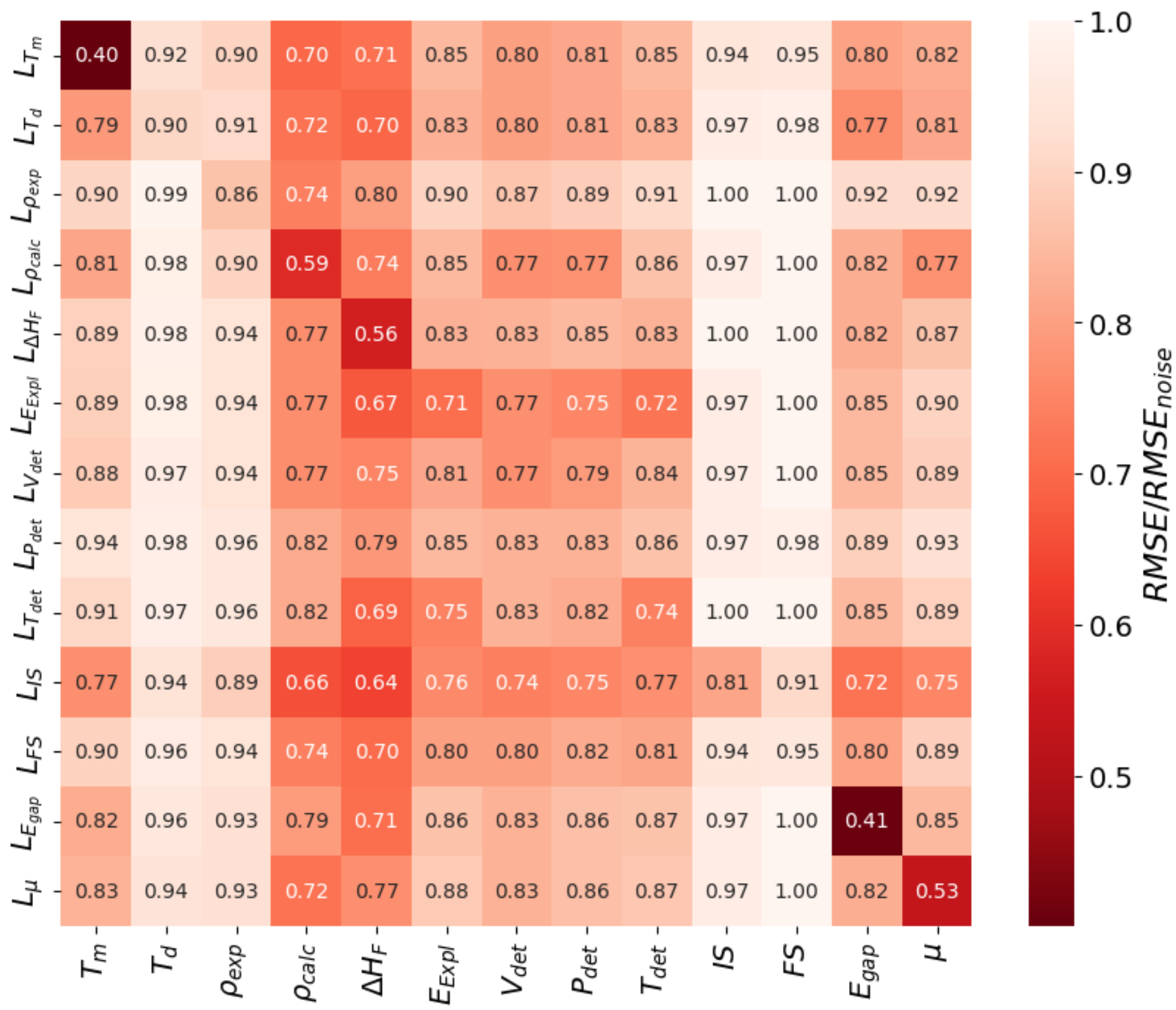

| | $T_m$ | $T_d$ | $\rho_{exp}$ | $\rho_{calc}$ | $\Delta H_F$ | $E_{Expl}$ | $V_{det}$ | $P_{det}$ | $T_{det}$ | $IS$ | $FS$ | $E_{gap}$ | $\mu$ |
|---|---|---|---|---|---|---|---|---|---|---|---|---|---|
| $L_{T_m}$ | 0.40 | 0.92 | 0.90 | 0.70 | 0.71 | 0.85 | 0.80 | 0.81 | 0.85 | 0.94 | 0.95 | 0.80 | 0.82 |
| $L_{T_d}$ | 0.79 | 0.90 | 0.91 | 0.72 | 0.70 | 0.83 | 0.80 | 0.81 | 0.83 | 0.97 | 0.98 | 0.77 | 0.81 |
| $L_{\rho_{exp}}$ | 0.90 | 0.99 | 0.86 | 0.74 | 0.80 | 0.90 | 0.87 | 0.89 | 0.91 | 1.00 | 1.00 | 0.92 | 0.92 |
| $L_{\rho_{calc}}$ | 0.81 | 0.98 | 0.90 | 0.59 | 0.74 | 0.85 | 0.77 | 0.77 | 0.86 | 0.97 | 1.00 | 0.82 | 0.77 |
| $L_{\Delta H_F}$ | 0.89 | 0.98 | 0.94 | 0.77 | 0.56 | 0.83 | 0.83 | 0.85 | 0.83 | 1.00 | 1.00 | 0.82 | 0.87 |
| $L_{E_{Expl}}$ | 0.89 | 0.98 | 0.94 | 0.77 | 0.67 | 0.71 | 0.77 | 0.75 | 0.72 | 0.97 | 1.00 | 0.85 | 0.90 |
| $L_{V_{det}}$ | 0.88 | 0.97 | 0.94 | 0.77 | 0.75 | 0.81 | 0.77 | 0.79 | 0.84 | 0.97 | 1.00 | 0.85 | 0.89 |
| $L_{P_{det}}$ | 0.94 | 0.98 | 0.96 | 0.82 | 0.79 | 0.85 | 0.83 | 0.83 | 0.86 | 0.97 | 0.98 | 0.89 | 0.93 |
| $L_{T_{det}}$ | 0.91 | 0.97 | 0.96 | 0.82 | 0.69 | 0.75 | 0.83 | 0.82 | 0.74 | 1.00 | 1.00 | 0.85 | 0.89 |
| $L_{IS}$ | 0.77 | 0.94 | 0.89 | 0.66 | 0.64 | 0.76 | 0.74 | 0.75 | 0.77 | 0.81 | 0.91 | 0.72 | 0.75 |
| $L_{FS}$ | 0.90 | 0.96 | 0.94 | 0.74 | 0.70 | 0.80 | 0.80 | 0.82 | 0.81 | 0.94 | 0.95 | 0.80 | 0.89 |
| $L_{E_{gap}}$ | 0.82 | 0.96 | 0.93 | 0.79 | 0.71 | 0.86 | 0.83 | 0.86 | 0.87 | 0.97 | 1.00 | 0.41 | 0.85 |
| $L_{\mu}$ | 0.83 | 0.94 | 0.93 | 0.72 | 0.77 | 0.88 | 0.83 | 0.86 | 0.87 | 0.97 | 1.00 | 0.82 | 0.53 |

**Figure 5.** The values for $a_{ij}$ after adding noise to each molecular embedding.

Inspecting Figure 5, we are not surprised that for most properties, the molecular embedding that most strongly affected the model predictions was the molecular embedding corresponding to the single-task model for that property. The exceptions to this are $V_{det}$, $P_{det}$, $T_{det}$, and FS. We also find that the model predictions for IS and FS are not strongly affected by the molecular embeddings of other properties. Also, we find that the molecular embedding for IS has a strong effect on the model predictions of most properties.

## Principal Component Analysis on Fused Molecular Embedding

In this section, we implement PCA as a data fusion technique for combining the molecular embeddings prior to training the F-MT model. We suspect that there are common features between different molecular embeddings that could be redundant when concatenated. By using the dimensionality reduction method of PCA on the concatenated molecular embedding we can reduce the size of the vector to n-components that maximize the variance of each feature in the molecular embedding across the entire training set. In each of the 5-folds, we fit PCA on the concatenated molecular embeddings across the training set for that fold and transform the molecular embeddings of the testing set for that fold. We explore PCA to compress the embeddings to a vector with 300 and 600 components. Table 5 contains the testing metrics for the F-MT models trained with the original concatenated molecular embedding and the three different PCA reduced molecular embeddings.

**Table 5.** Summary of test metrics across 5-fold cross validation comparing F-MT models with and without using PCA on the various properties of interest. The RMSE and $R^2$ values represent the mean across the 5-folds and the standard deviation is shown in parentheses.

| Test Predictions | | Models | | | | | |
|---|---|---|---|---|---|---|---|
| Subsets | Properties of Interest | F-MT D-MPNN (full) | | F-MT D-MPNN (300) | | F-MT D-MPNN (600) | |
| | | RMSE | $R^2$ | RMSE | $R^2$ | RMSE | $R^2$ |
| Thermal stability | $T_{melt}$ exp (K) | $36.7^{(1.6)}$ | $0.78^{(0.01)}$ | $36.7^{(1.6)}$ | $0.78^{(0.01)}$ | $37.2^{(1.2)}$ | $0.78^{(0.01)}$ |
| | $T_{dec}$ exp (K) | $49.3^{(4.2)}$ | $0.65^{(0.06)}$ | $49.7^{(4.6)}$ | $0.64^{(0.06)}$ | $49.4^{(4.6)}$ | $0.65^{(0.06)}$ |
| Crystal Properties | $\rho_0$ exp (g/cc) | $0.064^{(0.003)}$ | $0.86^{(0.01)}$ | $0.063^{(0.004)}$ | $0.86^{(0.02)}$ | $0.062^{(0.004)}$ | $0.87^{(0.02)}$ |
| | $\rho_0$ calc (g/cc) | $0.023^{(0.001)}$ | $0.98^{(0.01)}$ | $0.021^{(0.001)}$ | $0.98^{(0.01)}$ | $0.021^{(0.001)}$ | $0.98^{(0.01)}$ |
| | $\Delta H^0_f$ calc (kcal/mol) | $11.4^{(1.1)}$ | $0.99^{(0.01)}$ | $9.39^{(0.5)}$ | $0.99^{(0.01)}$ | $9.92^{(1.0)}$ | $0.99^{(0.01)}$ |
| Detonation Properties | $E_{expl}$ calc (kJ/cc) | $0.44^{(0.03)}$ | $0.97^{(0.01)}$ | $0.43^{(0.04)}$ | $0.97^{(0.01)}$ | $0.44^{(0.03)}$ | $0.97^{(0.01)}$ |
| | $V_{det}$ calc (km/s) | $0.20^{(0.01)}$ | $0.97^{(0.01)}$ | $0.20^{(0.01)}$ | $0.97^{(0.01)}$ | $0.20^{(0.01)}$ | $0.97^{(0.01)}$ |
| | $P_{det}$ calc (GPa) | $1.10^{(0.04)}$ | $0.97^{(0.01)}$ | $1.09^{(0.07)}$ | $0.97^{(0.01)}$ | $1.09^{(0.04)}$ | $0.97^{(0.01)}$ |
| | $T_{det}$ calc (K) | $136.9^{(7.6)}$ | $0.97^{(0.01)}$ | $132.9^{(10.4)}$ | $0.97^{(0.01)}$ | $134.4^{(9.6)}$ | $0.97^{(0.01)}$ |
| Sensitivity Properties | IS exp (J)* | $0.34^{(0.01)}$ | $0.58^{(0.03)}$ | $0.34^{(0.03)}$ | $0.56^{(0.08)}$ | $0.34^{(0.02)}$ | $0.56^{(0.07)}$ |
| | FS exp (N)* | $0.42^{(0.02)}$ | $0.47^{(0.11)}$ | $0.42^{(0.08)}$ | $0.47^{(0.14)}$ | $0.42^{(0.08)}$ | $0.47^{(0.11)}$ |
| Molecular Properties | $E_{gap}$ calc (eV) | $0.33^{(0.02)}$ | $0.91^{(0.01)}$ | $0.34^{(0.02)}$ | $0.90^{(0.01)}$ | $0.34^{(0.02)}$ | $0.91^{(0.01)}$ |
| | μ calc (Debye) | $1.67^{(0.04)}$ | $0.55^{(0.02)}$ | $1.67^{(0.03)}$ | $0.55^{(0.02)}$ | $1.67^{(0.04)}$ | $0.55^{(0.02)}$ |

* The values for these properties were converted with $\log_{10}$ for training and so the RMSE does not correspond to the real units.

The results suggest that generally the inclusion of PCA does not significantly improve the accuracy of the model, with exception of the properties $\rho_0$ calc and $\Delta H^0_f$ calc. We claim that the minimal effect of PCA on the predictability is because the FFN has the flexibility to transform the uncompressed embedding in a learned fashion to mimic any compression that PCA may provide. In terms of scalability, we find that PCA can be used to reduce the computational cost for training such models without a significant loss in accuracy. This could be particularly useful as the number of tasks increases to reduce the number of trainable parameters. In Figure 6, we plotted the number of trainable parameters for all the models explored in this work. We find that for both MT and F-MT models, the number of trainable parameters increases with the number of tasks. We note that the F-MT models increase at a much slower rate compared to the MT models, attributed to the fact that the F-MT models leverage pre-trained embeddings and do not need to learn the representations from scratch. Moreover, the F-MT models that were trained using the PCA reduced embeddings can dramatically reduce the number of trainable parameters even further.

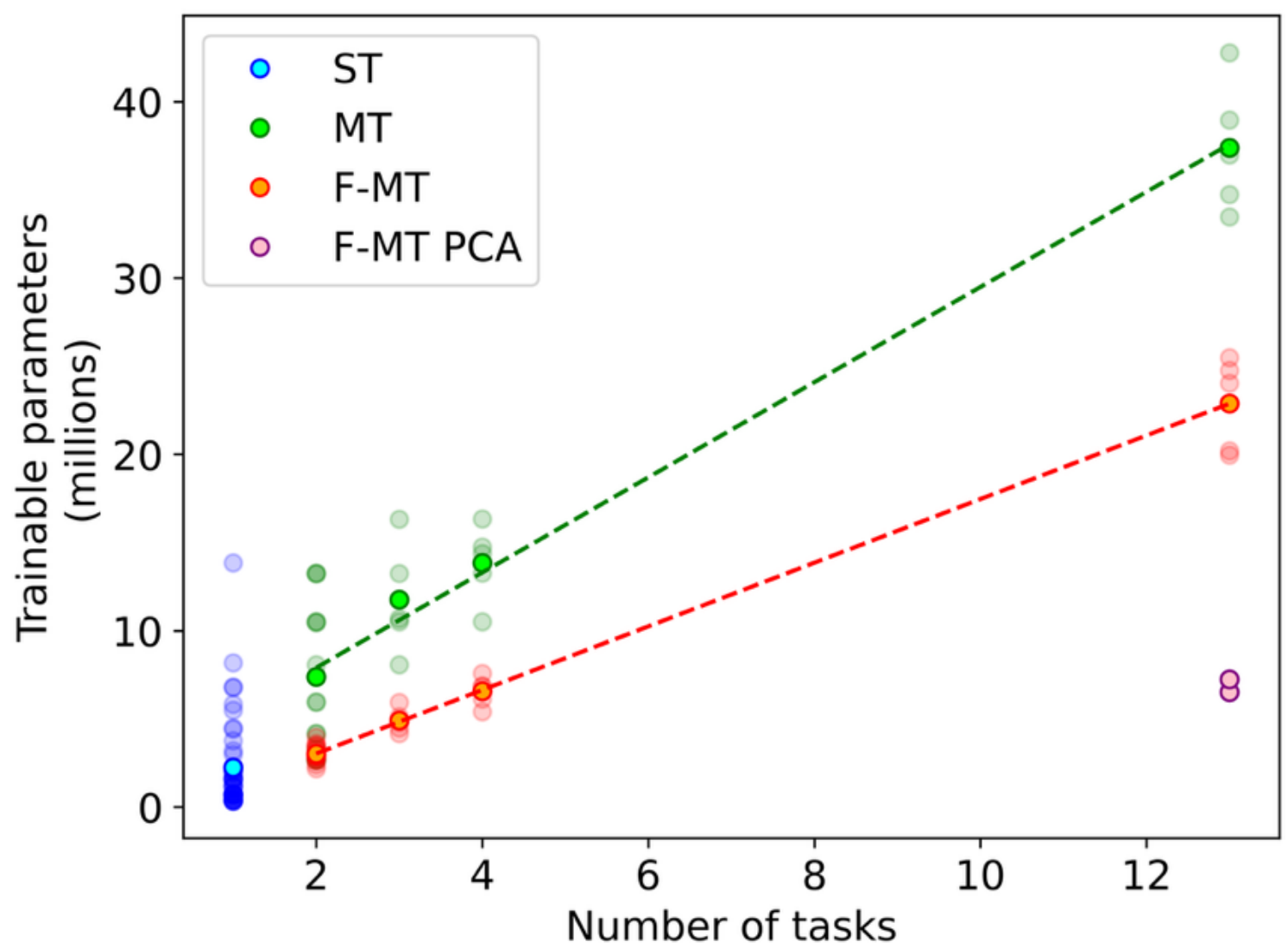

**Figure 6.** The number of trainable parameters for each model explored is shown in the transparent points. The solid points indicate the average for each combination of model type (ST, MT, or F-MT) and number of tasks. The dashed line indicates a linear fit to all the data for MT and F-MT models.

## Discussion

The goal of this work is to enhance the accuracy of MT modelling of D-MPNNs implemented in the *chemprop* framework. We designed and benchmarked a data fusion technique to combine latent representations of molecules from ST models such that the combined representation can be used to train a MT model. We apply F-MT models to a newly compiled dataset of ~30K unique molecules with 13 properties including experimental measurements collected from literature and properties obtain from our own quantum chemical and thermochemical calculations. The findings indicate that F-MT models not only outperform standard MT models on sparse datasets but also deliver improved predictive performance for data-limited properties compared to ST models. Surprisingly, our results on a complete dataset with strongly correlated properties indicate that F-MT models outperform standard ST and MT models, emphasizing the advantage of F-MT models over standard MT models and the robustness of this method. We demonstrate that F-MT models can provide deeper insights to the importance of and connections between the latent spaces of different properties compared to traditional methods such as CUR. The results of this work indicate a step forward in the ability to learn from sparse datasets that plague many areas of science. We believe that fusing deep-learned embeddings can enhance MT learning across diverse scientific fields, data types, and model architectures. For instance, in image classification one can combine embeddings obtained from convolutional neural networks with those from vision transformers to leverage complementary representations. Since training deep learning models can be computationally expensive, our approach offers a practical opportunity to leverage and extend the life cycle of existing pre-trained models. In standard MT modelling, to extend the model to predict a new task the entire model is trained from scratch, and the previous model is discarded. In the F-MT framework, only a single ST model needs to be trained on the data related to the new task while the deep learned embeddings of the previous tasks remain unchanged. Completing the F-MT model only requires training the FFN that maps the deep learned embeddings to each task. In addition, preliminary analysis using PCA on the fused embeddings suggests that dimensionality reduction techniques can effectively mitigate scalability issues as the number of regression tasks grows. Future work should study the saturation of the fused embedding as we suspect that the training of a new ST model may become unnecessary as the fused embedding becomes more descriptive.

The models presented in this work can efficiently predict fundamental properties relevant to EMs using only the molecular graph as input. This makes them well-suited for high-throughput screening of large molecular databases and for integration into generative design frameworks. A promising direction is the development of multi-objective inverse design approaches that learn to decode molecules from latent spaces trained on diverse property targets.

## Data and Software Availability

The newly compiled dataset and example code for training models on the QM9 dataset is available on GitHub https://github.itap.purdue.edu/StrachanGroup/FusedMultiTask. The chemprop software is available at https://github.com/chemprop.

## Acknowledgements

The authors thank Besty M. Rice, Edward F. C. Byrd, and Joshua L. Lansford for useful discussion and guidance.

This research study was sponsored by the Army Research Laboratory and was accomplished under Cooperative Agreement No. W911NF-20-2-0189. This work was supported in part by high-performance computer time and resources from the Department of Defense (DoD) High Performance Computing Modernization Program in collaboration with an appointment to the DoD Research Participation Program administered by the Oak Ridge Institute for Science and Education (ORISE) through an interagency agreement between the U.S. Department of Energy (DOE) and the DoD. ORISE is managed by ORAU under DOE contract number DE-SC0014664. All opinions expressed in this paper are the author's and do not necessarily reflect the policies and views of DoD, DOE, or ORAU/ORISE.

## Author Contributions

R.J.A. and B.C.B. conceptualized the work of this project. R.J.A. collected the data and developed the models. B.C.B. and A.S. supervised the work. B.C.B. developed the high-throughput workflow for the quantum chemistry and thermochemical calculations and provided HPC resources. All authors contributed to the interpretation of the results and the writing of the manuscript.

## Conflict of Interest

The authors declare no competing financial interests.